%% file: main.tex
\documentclass[runningheads]{style-eccv/llncs}

\usepackage{style-eccv/eccv}

\input{0-imports.tex}

\input{0-macros.tex}
\input{0-math.tex}

\usepackage{style-eccv/eccvabbrv}

\usepackage{graphicx}
\usepackage{booktabs}

\usepackage[accsupp]{axessibility}  % Improves PDF readability for those with disabilities.

\usepackage{orcidlink}

\begin{document}

\title{Audio-visual Generalized Zero-shot Learning \\ the Easy Way} 

\titlerunning{Audio-visual Generalized Zero-shot Learning the Easy Way}

\author{Shentong Mo\inst{1}$^*$\and Pedro Morgado\inst{2}}
\authorrunning{S. Mo and P. Morgado}

\institute{Carnegie Mellon University \and University of Wisconsin-Madison}

\maketitle

%%%%%%%%% ABSTRACT
\input{latex/0-abstract}

%%%%%%%%% BODY TEXT
\input{latex/1-intro}
\input{latex/2-related-work}

\input{latex/3-method}

\input{latex/4-experiments}
\input{latex/5-ablations}

\input{latex/6-conclusion}

\bibliographystyle{splncs04}
\bibliography{main}
\end{document}

%% file: 0-imports.tex
\usepackage[utf8]{inputenc} % allow utf-8 input
\usepackage[T1]{fontenc}    % use 8-bit T1 fonts
\usepackage{url}            % simple URL typesetting
\usepackage{nicefrac}       % compact symbols for 1/2, etc.
\usepackage{microtype}      % microtypography
\usepackage{booktabs}   % professional tables
\usepackage{tabularx}   % adjustable-width columns
\usepackage{colortbl}   % for coloring tables
\usepackage{multirow}
\usepackage{makecell}

\usepackage{wrapfig}
\usepackage{mwe} % for placeholder images

\usepackage[font=normal,labelfont=bf]{caption}
\usepackage[font=normal]{subcaption}

\usepackage[normalem]{ulem} % underlining
\usepackage{array}

\usepackage{xparse}
\usepackage{pifont}

\usepackage{ifthen}

\definecolor{cvprblue}{rgb}{0.21,0.49,0.74}
\usepackage[pagebackref,breaklinks,colorlinks,citecolor=cvprblue]{hyperref}

\usepackage[capitalize]{cleveref}
\crefformat{section}{Section~#2#1#3}
\crefformat{subsection}{Section~#2#1#3}
\crefformat{subsubsection}{Section~#2#1#3}
\crefformat{figure}{Fig.~#2#1#3}
\crefformat{equation}{Eq.~#2#1#3}
\crefformat{table}{Table~#2#1#3}
\crefformat{algorithm}{Alg.~#2#1#3}
\usepackage{algpseudocode}
\usepackage[linesnumbered,ruled,vlined]{algorithm2e}

\SetCommentSty{mycommfont}
\SetKwInput{KwInput}{Input}

\usepackage{enumitem}

\usepackage{xfp}

%% file: 0-macros.tex
\newcommand{\method}{{\fontfamily{lmtt}\selectfont\emph{\textsc{EZ-AVGZL}}}\xspace}
\newcommand{\methodFullName}{Easy Audio-Visual Generalized Zero-shot Learning\xspace}
\newcommand{\cmark}{\ding{51}}%
\newcommand{\xmark}{\ding{55}}%

\newlength\savewidth
\newlength\thinwidth

\definecolor{Gray}{gray}{0.93}
\newcolumntype{a}{>{\columncolor{Gray}}c}

\usepackage{xspace} % for abbreviation macros
\makeatletter
\DeclareRobustCommand\onedot{\futurelet\@let@token\@onedot}
\def\@onedot{\ifx\@let@token.\else.\null\fi\xspace}
 
\def\ie{\emph{i.e}\onedot}

\makeatother

%% file: 0-math.tex
\usepackage{amsmath,amsfonts,bm}

\def\1{\bm{1}}

\def\va{{\bm{a}}}

\def\vp{{\bm{p}}}

\def\vt{{\bm{t}}}

\def\vv{{\bm{v}}}
\def\vw{{\bm{w}}}
\def\vx{{\bm{x}}}

\DeclareMathAlphabet{\mathsfit}{\encodingdefault}{\sfdefault}{m}{sl}
\SetMathAlphabet{\mathsfit}{bold}{\encodingdefault}{\sfdefault}{bx}{n}

\def\gA{{\mathcal{A}}}

\def\gD{{\mathcal{D}}}

\def\gL{{\mathcal{L}}}

\def\gT{{\mathcal{T}}}

\def\gV{{\mathcal{V}}}

\def\gX{{\mathcal{X}}}

%% file: latex/0-abstract.tex
\begin{abstract}

Audio-visual generalized zero-shot learning is a rapidly advancing domain that seeks to understand the intricate relations between audio and visual cues within videos. 
The overarching goal is to leverage insights from seen classes to identify instances from previously unseen ones.
Prior approaches primarily utilized synchronized auto-encoders to reconstruct audio-visual attributes, which were informed by cross-attention transformers and projected text embeddings. 
However, these methods fell short of effectively capturing the intricate relationship between cross-modal features and class-label embeddings inherent in pre-trained language-aligned embeddings.
To circumvent these bottlenecks, we introduce a simple yet effective framework for \methodFullName, named \method, that aligns audio-visual embeddings with transformed text representations. 
It utilizes a single supervised text audio-visual contrastive loss to learn an alignment between audio-visual and textual modalities, moving away from the conventional approach of reconstructing cross-modal features and text embeddings.
Our key insight is that while class name embeddings are well aligned with language-based audio-visual features, they don't provide sufficient class separation to be useful for zero-shot learning. 
To address this, our method leverages differential optimization to transform class embeddings into a more discriminative space while preserving the semantic structure of language representations.
We conduct extensive experiments on VGGSound-GZSL, UCF-GZSL, and ActivityNet-GZSL benchmarks. Our results demonstrate that our \method achieves state-of-the-art performance in audio-visual generalized zero-shot learning. 

\keywords{Audio-Visual Learning \and Zero-shot Learning \and Audio-Visual Generalized Zero-shot Learning}

\end{abstract}

%% file: latex/1-intro.tex
\section{Introduction}

In the field of machine learning, the ability of models to recognize patterns and make predictions is crucial. Zero-shot learning exemplifies this capability, which leverages known information to understand the unknowns. To further explore its potential and applications, the principle of zero-shot learning has been applied in various contexts. One such context is Audio-visual generalized zero-shot learning (AVGZSL), an emerging topic of study that predicts unseen audio-visual categories by leveraging knowledge learned from seen classes.

Within the scope of AVGZSL, previous approaches~\cite{mercea2022avca,mercea2022tcaf} have focused on designing autodecoders for reconstructing audio-visual attributes. For example, TCaF~\cite{mercea2022tcaf} adopted a temporal contrastive feature alignment strategy. This strategy aligns audio-visual cues with word2vec embeddings of class prompts by defining both cross-entropy and regression losses to reduce the distance between the output embedding for a sample and the corresponding projected word2vec embedding. These cross-modal and textual features were derived from cross-attention transformers and the projected text embeddings. However, the complexity of the network for AVGZSL, which is derived from the synergy between cross-attention transformers and the corresponding projected word2vec embeddings, makes it challenging to understand the deeper interplay between cross-modal features and class label embeddings.

\begin{figure}[t]
\centering
\includegraphics[width=0.76\linewidth]{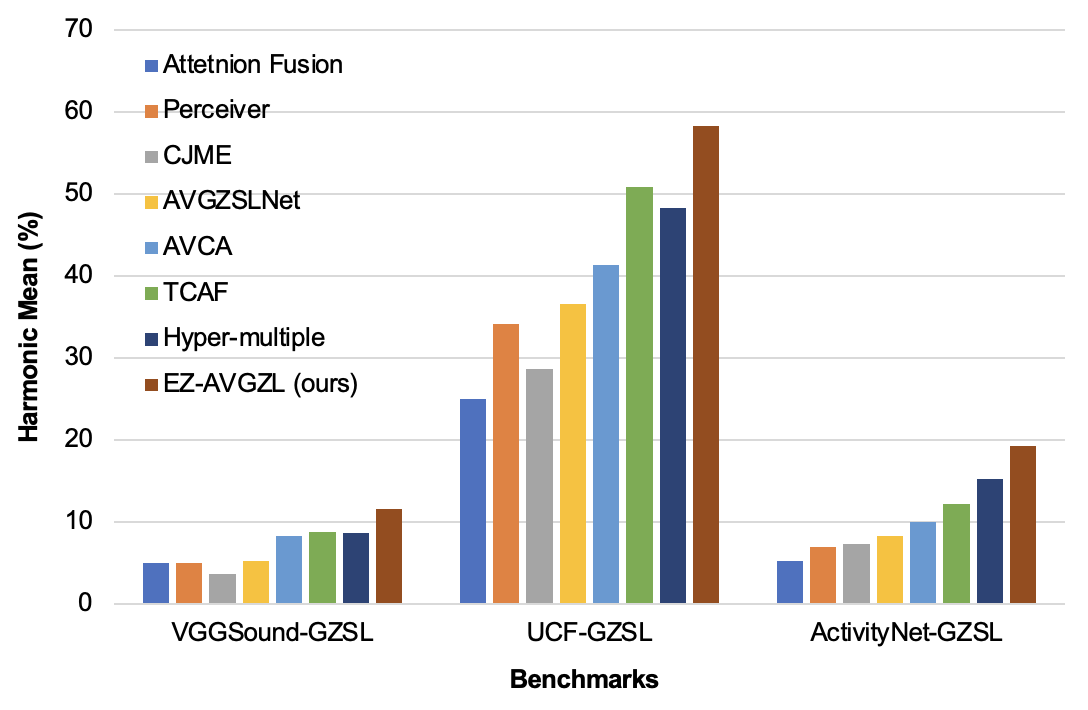}
\caption{Comparison of our \method with state-of-the-art methods on VGGSound-GZSL, UCF-GZSL, ActivityNet-GZSL benchmarks in terms of Harmonic Mean (higher is better) for seen and unseen classes. 
Our method significantly outperforms previous baselines in terms of all datasets.
}
\label{fig: title_img}
\end{figure}

In the context of audio-visual zero-shot learning, the main challenge lies in the absence of unseen classes during training. Without the supervision, the prediction of cross-modal class labels performs poorly on these audio-visual samples.
To address the challenge, we propose a novel framework for \methodFullName, referred to as \method. This framework aligns audio-visual embeddings with transformed text representations in a simple yet effective manner. \method learn well-separated text semantics across seen and unseen class names to achieve maximal separation while retaining the semantic relations between each other. During training, we leverage a simple supervised contrastive loss to learn the cross-modal alignment between audio-visual features and transformed text embeddings.
\method enhances class separability while preserving semantics through the use of differential optimization, thereby generating more distinct text embeddings.
Furthermore, it employs a single supervised text to audio-visual contrastive loss, establishing an alignment between the audio-visual and textual modalities. This approach effectively eliminates the need for reconstructing cross-modal features and text embeddings.

Empirical experiments on VGGSound-GZSL, UCF-GZSL, and ActivityNet-GZSL benchmarks demonstrate the state-of-the-art performance against previous audio-visual generalized zero-shot learning baselines, as illustrated in Fig.~\ref{fig: title_img}.
Extensive ablation studies further confirm the importance of class embedding optimization and supervised audio-visual language contrastive objectives in learning language alignment with audio-visual representations.
Additionally, we demonstrate the robustness of our approach across various zero-shot learning tasks and observe the benefit of our approach when applied to a variety of foundation models pre-trained from unimodal and cross-modal.

Our main contributions can be summarized as follows:
\begin{itemize}
    \item We introduce the \method framework, a paradigm shift in audio-visual generalized zero-shot learning. This framework simplifies the process with a supervised text-audio-visual contrastive loss, enhancing effectiveness.
    \item We propose a novel text transformation technique designed to construct distinct text embeddings, paving the way for better alignment and, by extension, superior recognition capabilities.
    \item We demonstrate that our method achieves state-of-the-art performance on extensive benchmarks in audio-visual generalized zero-shot learning.
\end{itemize}

%% file: latex/2-related-work.tex
\section{Related Work}

\noindent\textbf{Audio-Visual Learning.}
Audio-visual learning problems have been addressed in many previous works~\cite{aytar2016soundnet,owens2016ambient,Arandjelovic2017look,korbar2018cooperative,Senocak2018learning,zhao2018the,zhao2019the,Gan2020music,Morgado2020learning,Morgado2021robust,Morgado2021audio,hershey2001audio,ephrat2018looking,hu2019deep,mo2022semantic,mo2023diffava,mo2023oneavm,mo2024texttoaudio,mo2024semantic,zhang2024audiosynchronized} to learn the audio-visual correlation between two distinct modalities from videos.
Such cross-modal alignments are beneficial for a variety audio-visual tasks, including audio-event localization~\cite{tian2018ave,lin2019dual,wu2019dual,lin2020audiovisual,mo2022benchmarking,mo2022SLAVC,mo2023audiovisual,mo2023avsam,mo2023deepavfusion}, audio-visual spatialization~\cite{Morgado2018selfsupervised,gao20192.5D,Chen2020SoundSpacesAN,Morgado2020learning}, audio-visual continual learning~\cite{pian2023audiovisual,mo2023classincremental}, audio-visual navigation~\cite{Chen2020SoundSpacesAN,chen2021waypoints,chen22soundspaces2} and audio-visual parsing~\cite{tian2020avvp,wu2021explore,lin2021exploring,mo2022multimodal}.
In this work, our primary focus is on learning audio-visual representations that align with text embeddings for generalized zero-shot learning, a task that presents more challenges than the aforementioned tasks.

\noindent\textbf{Zero-shot Learning.}
Zero-shot learning aims to leverage semantic relationships, typically utilizing attributes or word embeddings associated with class labels to make informed predictions about previously unseen categories. This is often achieved by drawing parallels to the semantic information of known categories.
Early audio-visual zero-shot learning methods, such as those proposed by Parida et al.\cite{Parida2020cjme} and Mazumder et al.\cite{Mazumder2021avgzslnet}, attempted to map video, audio, and text into the same feature space and compute the similarity across each other for comparison.
For example, Audio-Visual Generalized Zero-shot Learning Network~\cite{Mazumder2021avgzslnet} introduced a reconstruction module to predict text representations from audio and visual features.
More recently, ImageBind~\cite{girdhar2023imagebind} proposed learning a joint embedding across six different modalities, especially on audio, text, and video. This approach demonstrated the ability to recognize or retrieve unseen categories without explicit training.
Unlike these methods, our approach does not require large-scale pre-training.
Instead, we present a simple yet effective approach based on unimodal or crossmodal weights pre-trained on large-scale data to achieve audio-visual generalized zero-shot learning.

\noindent\textbf{Audio-Visual Generalized Zero-shot Learning.}
Audio-visual generalized zero-shot learning extends the zero-shot learning paradigm to scenarios where both seen and unseen classes are present at test time. 
The seminal work, AVCA~\cite{mercea2022avca} proposed an attention mechanism to selectively focus on discriminative audio and visual cues. 
By harnessing the combined strength of these cues, AVCA not only achieved superior performance on seen classes but also exhibited proficiency in recognizing unseen classes.

Following up, TCaF\cite{mercea2022tcaf} introduced a temporal contrastive feature alignment strategy. This strategy aligns audio-visual cues across time and applies two separate decoders to reconstruct the audio-visual embedding and the projected word2vec embedding to be close to the original word2vec embedding of class labels.
More recently, Hong {\it et al.}~\cite{hong2023hyperbolic} incorporated cross-modality alignment between video and audio features in the hyperbolic space. They used multiple adaptive curvatures for hyperbolic projections to achieve curvature-aware geometric learning.
In contrast, we develop a novel framework to aggregate the alignment between audio-visual representations and text embeddings with explicit class embedding optimization.
To the best of our knowledge, this work is the first to leverage an explicit class embedding optimization mechanism for audio-visual generalized zero-shot learning.
Our experiments in Section~\ref{sec:exp} also demonstrate the effectiveness of our method in classifying both seen and unseen categories.

%% file: latex/3-method.tex
\section{Method}
We aim to learn a language-aligned representation for a video clip, which includes both the visual frame sequence and corresponding audio, to enable open-dictionary recognition of audio-visual events. To achieve this, we propose a simple yet effective approach, named \methodFullName (\method). This approach seeks to 1) construct multi-modal representations of the data and 2) align them with language representations using a supervised contrastive learning objective while leveraging strong foundation models. The proposed approach is based on two components: class embedding optimization described in Section~\ref{sec:tt} and supervised alignment between language and audio-visual representations described in Section~\ref{sec:stavc}. Despite its simplicity, \method outperforms recent state-of-the-art methods on Audio-Visual Generalized Zero-shot Learning (AVGZSL) on a variety of datasets.

\begin{figure*}[t]
\centering
\includegraphics[width=0.95\linewidth]{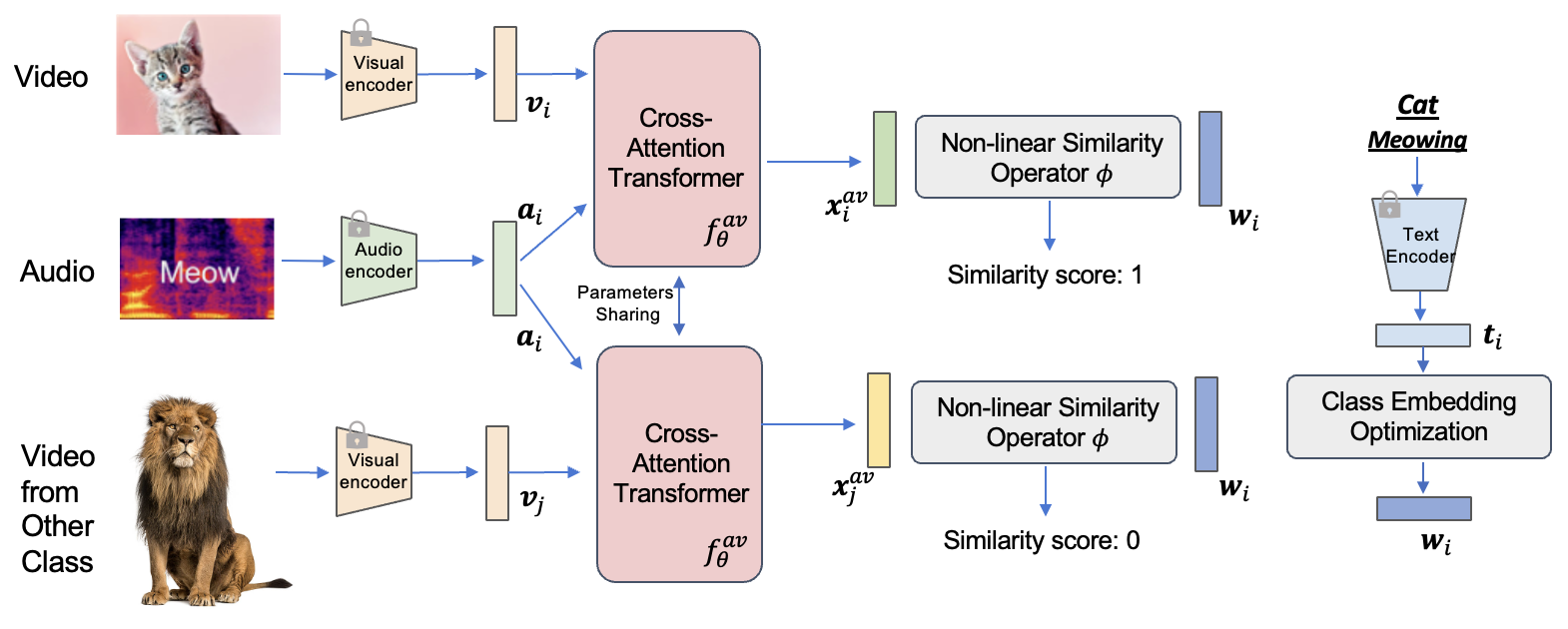}
\caption{Illustration of the proposed Easy Audio-Visual Generalized Zero-shot Learning (\method) framework. 
The initial class embbedings $\vt_i$ from a frozen text encoder are optimized with maximal separability and preserved semantics to generate the new embeddings $\vw_i$.
Then the cross-attention transformer $f_\theta^{av}(\cdot,\cdot)$ takes visual and audio features $(\vv_i, \va_i)$ from the unimodal encoder to generate the multi-modal representations $\vx^{av}_i$. 
Finally, a non-linear similarity function is applied to align representations $\vx^{av}_i$ with the corresponding class embeddings $\vw_{y_i}$ by minimizing the cross-entropy loss between the predicted similarity score and the target score as 1 in the target class entry, while the target is 0 for multi-modal representations $\vx^{av}_j$ given visual features $\vv_j$ of videos from other classes.
}
\label{fig: main_img}
\end{figure*}

\subsection{Preliminaries: Audio-Visual Generalized Zero-shot Learning (AVGZSL)}
% We start by introducing AVGZSL problem. 
\noindent\textbf{Problem Statement.}
The objective of AVGZSL is to utilize natural language descriptions of audio-visual events $t_i$ and the semantic relationships to enable a model to recognize events from a set of audio-visual event classes $\gT=\gT_s\cup\gT_u$. This should be possible regardless of whether the model has seen them ($t_i\in\gT_s$) or not seen them ($t_i\in\gT_u$) during training.

To accomplish this, a model $f_\theta:\gV\times\gA\rightarrow\gT$ is trained on the seen dataset $\gD_s=\{(v_i, a_i, t_i)\}$ for audio-visual event recognition, where $v_i\in\gV$ denotes the $i^{th}$ visual clip, typically multiple frames spanning a short time window, $a_i\in\gA$ denotes the corresponding audio, and $t_i\in\gT_s$ denotes the (human-readable) class name of the audio-visual event depicted in video $(v_i, a_i)$.
For effective zero-shot learning, a model needs to learn a mapping that can generalize beyond $\gT_s$ by inferring from the semantic relationships between seen and unseen classes. In other words, the AVGZSL setting evaluates models based on their classification performance on videos $(v_i, a_i)$ belonging to both seen $\gT_s$ and unseen classes $\gT_u$.

\noindent\textbf{Architecture}
Following TCaF~\cite{mercea2022tcaf}, we tackle AVGZSL using a variety of pre-trained models to encode all input signals. We experiment with a wider variety of pre-trained models than in~\cite{mercea2022tcaf}, in all cases, we train our system on a series of unimodal pre-extracted feature representations for the visual frames $\vv_i=f_v(v_i)$, audio $\va_i=f_a(a_i)$ and class names $\vt_i=f_t(t_i)$. As our goal is multi-modal GZSL, we further encode the unimodal audio and visual representations into a joint multi-modal representation using the fusion model $\vx_i^{av} = f_{av}(\vv_i, \va_i)$ proposed in~\cite{mercea2022tcaf}, which consists of a sequence of $L$ transformer layers with cross-attention between audio and visual modalities. 

The quality of representations obtained from foundation models is crucial for effective zero-shot learning. However, the focus of this work is the language alignment step. While prior work often relies on elaborate alignment strategies, we demonstrate that a simple supervised contrastive objective between text and audio-visual representations can achieve state-of-the-art performance. This is possible as long as the class name embeddings for seen and unseen events are maximally separated while retaining the semantic relations between each other. An overview of the proposed approach is shown in \cref{fig: main_img}.

Apart from the quality of representations derived from foundation models, language alignment, which is the focus of this work, is a critical component for effective zero-shot learning. 
While prior work often relies on complex alignment strategies, we demonstrate that a \textit{\textbf{simple supervised contrastive objective}} between text and audio-visual representations can achieve state-of-the-art performance, as long as the class name embeddings for seen and unseen events are \textit{\textbf{maximally separated}} while retaining the \textit{\textbf{semantic relations}} between each other.
An overview of the proposed approach is shown in \cref{fig: main_img}.

\subsection{Class Embedding Optimization}
\label{sec:tt}

Language-aligned classifiers can achieve zero-shot generalization as class embeddings $\vt_1,\ldots,\vt_{|\gT|}$ obtained from large-scale models like LLaMa or CLIP, encode semantic relationships between classes. However, while these class embeddings can effectively encode semantics, they are sub-optimal for classification due to a lack of class separability. To obtain well-separated semantic embeddings, we optimize a new set of class embeddings $\vw^*_1,\ldots,\vw^*_{|\gT|}$ to satisfy two competing constraints: class separability and semantic preservation.

\noindent\textbf{Class Separability.}
Class embeddings that are maximally separated from each other are better suited for recognition of seen classes~\cite{MorgadoProxyHashing}. Thus, to achieve class embedding separation, we minimize the negative distance of each class embedding to its nearest neighbor
\begin{equation}
    \label{eq:sep}
    \mathcal{L}_{sep} = -\sum_{c=1}^{|\gT|} \min_{k \neq c}\|\vw_c -\vw_k\|_2.
\end{equation}
However, maximizing separability alone would destroy the semantics of the pre-trained embeddings. In fact, the minimum of~\cref{eq:sep} (if norm constrained) would be achieved for a set of embeddings $\vw_1,\ldots,\vw_{|\gT|}$ following a "Tammes problem" configuration~\cite{tammes1930origin}. 
Note that the embeddings are normalized. 
When the embedding dimension $d$ is larger than the number of classes in the dictionary $\gT$, the Tammes configuration provides a solution where all embeddings are equally spaced from each other, \ie ${\|\vw_i-\vw_j\|_2=\text{Cte}\quad \forall i \neq j}$. For example, the distance between the class "a person playing guitar" and the class "a person playing the violin" would be the same as the distance to the class "a dog barking". 
Hence, since effective zero-shot recognition heavily relies on the semantic relations between seen and unseen classes, class separability must be balanced with the preservation of the semantics available in pre-trained embeddings.

\noindent\textbf{Semantic preservation}
To preserve the semantics of the initial embeddings $\vt_i$, a simple approach would be to use $\vt_i$ as proxies and constrain the new embeddings $\vw_i$ to be in their proximity
\begin{equation}
    \label{eq:sem_prox}
    \mathcal{L}^{\textit{prox}}_{sem} = \sum_{c=1}^{|\gT|}\|\vw_c -\vt_c\|_2.
\end{equation}
However, this constraint would significantly limit the separability of class embeddings $\vw$, given their direct ties to $\vt$. Instead, we seek new class embeddings $\vw$ that retain a similar ranking of similarities to other classes. This is accomplished using a margin ranking loss of the form
\begin{equation}\label{eq:sem}
     \mathcal{L}_{sem}^{\textit{rank}} = \underset{c \neq i \neq j}{\sum_{c=1}^{|\gT|}\sum_{i=1}^{|\gT|}\sum_{j=1}^{|\gT|}} 
     % \sum_{\substack{c,i,j=1\\c \neq i \neq j}}^{|\gT|} 
     \max\left\{0, m-\text{sign}\left(d^\vt_{ci} - d^\vt_{cj}\right) \left(d^\vw_{ci} - d^\vw_{cj}\right)\right\}
\end{equation}
where $m$ is a margin coefficient, and $d^\vt_{ci}=\|\vt_c-\vt_i\|_2$ and $d^\vw_{ci}=\|\vw_c-\vw_i\|_2$ are the $\ell_2$ distances between pairs of class embeddings, using either the pre-trained embeddings $\vt$ or the new optimized embeddings $\vw$, respectively. Intuitively, this ranking loss will seek to organize the new embeddings $\vw$ such that if $\vt_c$ is closer to $\vt_i$ than to $\vt_j$, then $\vw_c$ will also be closer to $\vw_i$ than to $\vw_j$.
Note that Eq.~\ref{eq:sem}'s complexity grows cubically with the number of classes. 
While manageable for the tested datasets, it can be impractical for large vocabularies. 
A potential solution is to sample a large (but fixed) number of class triplets per iteration, and replace Eq.~\ref{eq:sem} with an expectation over these triplets.

To balance separability and semantic preservation, we obtain the class embeddings by minimizing the joint loss 
\begin{equation}
    \label{eq:tt}
    \underset{\vw_1,\ldots,\vw_{|\gT|}}{\arg\min} (1 - \alpha) \gL_{sem} + \alpha \gL_{sep},
\end{equation}
where $\alpha$ is set to 0.5. Given that the objective is differentiable, we optimize the new embeddings using gradient descent, starting from the initialization $\vw_i=\vt_i\,\forall i$.

\subsection{Supervised Audio-Visual Language Alignment}
\label{sec:stavc}
Given the enhanced class embeddings $\vw_1,\ldots,\vw_{|\gT|}$ (with improved class separability), we empirically show that a contrastive objective is sufficient to learn state-of-the-art models for generalized zero-shot learning of audio-visual events. Specifically, let $\vx^{av}_i=f_\theta^{av}(\vv_i, \va_i)$ be the multi-modal representation of an audio-visual sample $(v_i, a_i)$ belonging to class $y_i$. 
Then, we align representations $\vx^{av}_i$ with the corresponding class embeddings $\vw_{y_i}$ by minimizing the contrastive loss
\begin{equation}
    \label{eq:supcon}
    \mathcal{L}_{\textit{AVLA}} = 
    - \log
    \dfrac{
    \exp\left(\text{sim}(\vx^{av}_i, \vw_{y_i})\right)}{
    \sum_{k=1}^B\exp(\text{sim}(\vx^{av}_i, \vw_{y_k}))}
\end{equation}
where $\text{sim}(\cdot, \cdot)$ is a similarity function used to contrast the class embedding $\vw_{y_i}$ against others in the current batch of data $\vw_{y_k}$.

\noindent\textbf{Similarity.} While contrastive objectives typically rely on cosine similarity metrics for $\text{sim}(\cdot,\cdot):\gX\times\gT\rightarrow\Re$, we found that non-linear similarity functions can better capture the relationship between audio-visual representations and class embeddings. 
Specifically, we draw inspiration from transformer attention modules and use a multi-head cross-attention layer. In this layer, class embeddings $\vw_c$ are used as queries and audio-visual representations $\vx^{av}_i$ as key-value pairs. 
After cross-attention, class similarities are computed through a linear projection $\vp$.
Optimizing \cref{eq:supcon} pushes the learned embeddings closer to the text embeddings of the same category. During inference, we use the cross-attention transformer layer to output $C$ similarity scores corresponding to each category, including unseen ones. We then classify the audio-visual samples according to the class with highest similarity.

%% file: latex/4-experiments.tex
\section{Experiments}
In this section, we evaluate the proposed \method on three benchmarks for audio-visual generalized zero-shot learning, including VGGSound-GZSL~\cite{mercea2022avca}, UCF-GZSL~\cite{mercea2022avca}, and ActivityNet-GZSL~\cite{mercea2022avca}.

\subsection{Experimental Setup}

\noindent\textbf{Datasets.}
VGGSound-GZSL~\cite{mercea2022avca} contains 93,752 videos across 276 classes, selected from VGGSound~\cite{chen2020vggsound}, a large audio-visual dataset of 309 classes over 200k videos.
We follow the prior work~\cite{mercea2022avca,mercea2022tcaf} and use the same split for train/val(U)/test(U) over 138/69/69 classes.
UCF-GZSL~\cite{mercea2022avca} includes 6,816 videos from 51 classes, chosen from UCF101~\cite{soomro2012ucf101}, the original video action recognition dataset, which consists of over 13k videos in 101 classes.
We use the same train/val(U)/test(U) split over 30/12/9 classes in previous work~\cite{mercea2022avca,mercea2022tcaf}.
ActivityNet-GZSL~\cite{mercea2022avca} contains 20k videos in 200 classes of varying duration for action recognition~\cite{heilbron2015activitynet}.
We use the same train/val(U)/test(U) split over 99/51/50 classes as in the prior work~\cite{mercea2022avca,mercea2022tcaf} 

\noindent\textbf{Evaluation Metrics.}
In line with prior work~\cite{mercea2022avca,mercea2022tcaf}, we use the mean class accuracy to evaluate all models. 
For ZSL performance, we evaluate only on the subset of test samples from unseen classes. 
For the GZSL performance, we evaluate on the full test set, which includes both seen and unseen classes. 
We report results for the subsets of seen (S) and unseen (U) classes, as well as their harmonic mean.

\noindent\textbf{Implementation.}
We follow previous work~\cite{mercea2022tcaf} and use 8 heads with a dimension of 64 per head for all attention blocks in the cross-attention transformer. 
The model is trained for 50 epochs with a batch size of 64, using the Adam optimizer~\cite{kingma2014adam} with a learning rate of 1e-4, running average coefficients $\beta_1$ = 0.9, $\beta_2$ = 0.999, and a weight decay of 1e-5. 
Following the prior work~\cite{mercea2022tcaf,mercea2022avca}, we use the word2vec~\cite{mikolov2013efficient} embeddings for text, and VGGish~\cite{hershey2017cnn} embeddings for audio, and C3D~\cite{tran2015learning} features for video frames.

\input{tables/exp_vggsound}

\input{tables/exp_ucf}

\input{tables/exp_activitynet}

\subsection{Comparison to Prior Work}
\label{sec:exp}

In this work, we propose a novel and effective framework for audio-visual generalized zero-shot learning.
In order to demonstrate the effectiveness of the proposed \method, we comprehensively compare it to previous audio-visual baselines~\cite{Fayek2021large,jaegle2021perceiver} adapted to our problem and current state-of-the-art audio-visual generalized zero-shot learning approaches~\cite{Parida2020cjme,Mazumder2021avgzslnet,mercea2022avca,mercea2022tcaf,hong2023hyperbolic}. 

For the VGGSound-GZSL dataset, we report the quantitative comparison results in Table~\ref{tab: exp_vggsound}.
As shown, we achieve the best results in all metrics for both seen and unseen classes compared to previous audio-visual generalized zero-shot learning approaches.
In particular, the proposed \method outperforms Hyper-multiple~\cite{hong2023hyperbolic}, the current state-of-the-art audio-visual generalized zero-shot learning baseline, by 1.64@Seen, 2.68@Unseen, 2.88@Harmonic Mean, and 1.61@ZSL.
Moreover, we also perform better than TCaF~\cite{mercea2022tcaf}, a strong audio-visual generalized zero-shot learning baseline, by 4.63@Seen, 1.96@Unseen, 2.78@Harmonic Mean, and 1.51@ZSL.
Furthermore, we achieve significant performance gains compared to AVCA~\cite{mercea2022avca}, the first baseline using cross-modal attention and textual label embeddings. This indicates the importance of optimized text transformation and non-linear similarity functions in learning better transferrable knowledge for generalized zero-shot learning.

Additionally, we observe significant gains in the UCF-GZSL and ActivityNet-GZSL benchmarks, as shown in Table~\ref{tab: exp_ucf} and Table~\ref{tab: exp_activitynet}. 
Compared to TCaF~\cite{mercea2022tcaf}, a strong baseline using three manually-designed objectives, we achieve result gains of 11.18@Seen, 5.52@Unseen, 7.46@Harmonic Mean, and 3.63@ZSL on UCF-GZSL dataset.
When evaluated on the challenging ActivityNet-GZSL benchmark, our method still outperforms TCaF~\cite{mercea2022tcaf} by 8.40@Seen, 5.20@Unseen, 7.07@Harmonic Mean, and 5.05@ZSL.
We also achieve better results than AVCA~\cite{mercea2022avca}, a cross-modal baseline based on cross-modal attention between audio and visual embeddings.
These results demonstrate the effectiveness of our approach in learning optimized textual label embeddings and non-linear similarity functions for generalized audio-visual zero-shot recognition.

%% file: tables/exp_vggsound.tex
\begin{table}[t]
	%\normalem
	\renewcommand\tabcolsep{3.0pt}
	\centering
        \caption{Quantitative results of VGGSound-GZSL benchmark.}
   \label{tab: exp_vggsound}
	\resizebox{0.6\linewidth}{!}{
		\begin{tabular}{lcccc}
			\toprule
    \bf Method & \bf Seen & \bf Unseen & \bf Harmonic Mean & \bf ZSL \\
   \midrule
    AttentionFusion~\cite{Fayek2021large} & 14.13 & 3.00 & 4.95 & 3.37 \\
    Perceiver~\cite{jaegle2021perceiver} & 13.25 & 3.03 & 4.93 & 3.44 \\
    CJME~\cite{Parida2020cjme} & 10.86 & 2.22 & 3.68 & 3.72 \\
    AVGZSLNet~\cite{Mazumder2021avgzslnet} & 15.02 & 3.19 & 5.26 & 4.81 \\
    AVCA~\cite{mercea2022avca} & 12.63 & 6.19 & 8.31 & 6.91 \\
    TCaF~\cite{mercea2022tcaf} & 12.63 & 6.72 & 8.77 & 7.41 \\
    Hyper-alignment~\cite{hong2023hyperbolic} & 12.50	& 6.44 & 8.50 & 7.25 \\
    Hyper-single~\cite{hong2023hyperbolic} & 12.56	& 5.03 & 7.18 & 5.47 \\
    Hyper-multiple~\cite{hong2023hyperbolic} & 15.62	& 6.00 & 8.67 & 7.31 \\
    \rowcolor{blue!10}
    \method (ours) & \bf 17.26	& \bf 8.68 & \bf 11.55 & \bf 8.92 \\
   \bottomrule
	\end{tabular}}
\end{table}

%% file: tables/exp_ucf.tex
\begin{table}[t]
	%\normalem
	\renewcommand\tabcolsep{3.0pt}
	\centering
        \caption{Quantitative results of UCF-GZSL benchmark.}
   \label{tab: exp_ucf}
	\resizebox{0.6\linewidth}{!}{
		\begin{tabular}{lcccc}
			\toprule
    \bf Method & \bf Seen & \bf Unseen & \bf Harmonic Mean & \bf ZSL \\
   \midrule
    AttentionFusion~\cite{Fayek2021large} & 39.34	& 18.29 & 24.97 & 20.21 \\
    Perceiver~\cite{jaegle2021perceiver} & 46.85	& 26.82 & 34.11 & 28.12 \\
    CJME~\cite{Parida2020cjme} & 33.89	& 24.82 & 28.65 & 29.01 \\
    AVGZSLNet~\cite{Mazumder2021avgzslnet} & 74.79	& 24.15 & 36.51 & 31.51 \\
    AVCA~\cite{mercea2022avca} & 63.15	& 30.72 & 41.34 & 37.72 \\
    TCaF~\cite{mercea2022tcaf} & 67.14	& 40.83 & 50.78 & 44.64 \\
    Hyper-alignment~\cite{hong2023hyperbolic} & 57.13	& 33.86	& 42.52	& 39.80 \\
    Hyper-single~\cite{hong2023hyperbolic}  & 63.47	& 34.85	& 44.99	& 39.86 \\
    Hyper-multiple~\cite{hong2023hyperbolic} & 74.26	& 35.79	& 48.30	& \bf 52.11 \\
    \rowcolor{blue!10}
    \method (ours) & \bf 78.32	& \bf 46.35 & \bf 58.24 & 48.27 \\
   \bottomrule
	\end{tabular}}
\end{table}

%% file: tables/exp_activitynet.tex
\begin{table}[t]
	%\normalem
	\renewcommand\tabcolsep{3.0pt}
	\centering
        \caption{Quantitative results of ActivityNet--GZSL benchmark.}
   \label{tab: exp_activitynet}
	\resizebox{0.6\linewidth}{!}{
		\begin{tabular}{lcccc}
			\toprule
    \bf Method & \bf Seen & \bf Unseen & \bf Harmonic Mean & \bf ZSL \\
   \midrule
    AttentionFusion~\cite{Fayek2021large} & 11.15 & 3.37	 &5.18	  & 4.88 \\
    Perceiver~\cite{jaegle2021perceiver} & 18.25 & 4.27	 &6.92	  & 4.47 \\
    CJME~\cite{Parida2020cjme} & 10.75 & 5.55	 &7.32	  & 6.29 \\
    AVGZSLNet~\cite{Mazumder2021avgzslnet} & 13.70 & 5.96	 &8.30	  & 6.39 \\
    AVCA~\cite{mercea2022avca} & 16.77 & 7.04	 &9.92	  & 7.58 \\
    TCaF~\cite{mercea2022tcaf} & 30.12 & 7.65	 &12.20  & 7.96 \\
    Hyper-alignment~\cite{hong2023hyperbolic} & 29.77	& 8.77 & 13.55 & 9.13 \\
    Hyper-single~\cite{hong2023hyperbolic}  & 24.61	& 10.10 & 14.32 & 10.37 \\
    Hyper-multiple~\cite{hong2023hyperbolic} & 36.98	& 9.60 & 15.25 & 10.39 \\
    \rowcolor{blue!10}
    \method (ours) & \bf 38.52 & \bf 12.85 & \bf 19.27  & \bf 13.01 \\
   \bottomrule
	\end{tabular}}
\end{table}

%% file: latex/5-ablations.tex
\subsection{Experimental Analysis}\label{sec:exp_analysis}

In this section, we conduct ablation studies to demonstrate the benefit of introducing the Class Embedding Optimization (CEO) and Audio-Visual Language Alignment (AVLA) modules.
We also carry out extensive experiments to investigate optimization objectives in class embedding optimization, non-linear similarity functions, and generalization across diverse cross-modal settings.

\input{tables/ab_component}

\noindent\textbf{Class Embedding Optimization (CEO) \& Audio-Visual Language Alignment (AVLA).}
In order to demonstrate the effectiveness of CEO and AVLA, we conduct an ablation study and report the quantitative results on VGGSound-GZSL dataset in Table~\ref{tab: exp_ablation}.

As observed, adding CEO to the vanilla baseline improves the results by 2.72@Seen, 1.20@Unseen, 1.68@Harmonic Mean, and 0.95@ZSL, which validates the benefit of optimizing class embeddings so as to obtain separated embeddings that preserve the semantic relationships between classes.
On the other hand, introducing AVLA in the baseline increases the zero-shot performance across all metrics.
More importantly, incorporating both CEP and AVLA into the baseline significantly raises the performance by 4.63@Seen, 1.96@Unseen, 2.78@Harmonic Mean, and 1.51@ZSL.
These improvements validate the importance of text transformation and supervised text audio-visual contrastive in extracting separated textual label embeddings for AVGZSL.

\noindent\textbf{Objectives in Class Embedding Optimization.}
Optimization objectives in class embedding transformation, including global semantic preservation and class separability, are essential for learning optimized textual class embeddings for zero-shot learning.
To explore the effects of each objective more comprehensively, we ablate each text optimization objective and report the quantitative results in Table~\ref{tab: ab_textloss}.
We observe that adding the semantic preservation loss to the vanilla baseline improves the performance by 1.08@Seen, 0.60@Unseen, 1.14@Harmonic Mean, and 0.35@ZSL.
On the other hand, introducing the class separability loss in the baseline enhaces the zero-shot performance across all metrics.
Furthermore, combining the semantic preservation and class separability objectives for the text transformer significantly increases the performance by 2.59@Seen, 3.05@Unseen, 3.78@Harmonic Mean, and 2.45@ZSL. This result validates the importance of class separability loss in balancing the preservation of the semantics available in pre-trained representations for zero-shot learning.
Replacing the Euclidean-based semantic preservation loss $\mathcal{L}^{\textit{prox}}_{sem}$ with a ranking based loss $\mathcal{L}^{\textit{rank}}_{sem}$ further achieves performance gains of 0.79@Seen, 1.60@Unseen, 1.55@Harmonic Mean, and 1.44@ZSL.
These results demonstrate the effectiveness of both semantic preservation and class separability objectives in class embedding optimization to generate meaningful text representations.

\noindent\textbf{Non-linear Similarity Functions.}
The non-linear similarity operator used in supervised audio-visual language alignment impacts the extracted cross-modal representations.
To explore these effects more comprehensively, we varied the functions from $\{$Cosine-similarity, Linear, MLP, Cross-Attention$\}$.
For Cosine-Similarity, we directly compute the dot product between audio-visual embeddings $\mathbf{x}^{av}_i$ and text embeddings $\mathbf{t}_i$.
For Linear, we apply a linear layer to the concatenation of both embeddings to predict the similarity score.
For MLP, we replace the linear layer with a multi-layer perceptron to generate the score.
For Cross-Attention, we use text embeddings $\mathbf{t}_i$ as the queries and audio-visual representations $\mathbf{x}^{av}_i$ as key-value pairs in a transformer attention module to output the score.
The comparison results of zero-shot performance on VGGSound-GZSL benchmark are reported in Table~\ref{tab: ab_simfunc}.
When using Cosine-Similarity as the function, we achieve the worst results across all metrics.
Replacing the cosine-similarity operator with both Linear increases the seen results by 2.55@Seen, but decreases the performance by 0.07@Unseen and 0.12@ZSL.
This might be because adding complexity to the function causes the model to overfit the seen classes in the training data, , leading to poorer generalization to unseen categories.
Additionally, using MLP as the function operator significantly raises the performance by 3.06@Seen, but slightly deteriorates the results on unseen classes.
Our method achieves the best results across all metrics by using the cross-attention operator to predict the similarity score.
These improved results demonstrate the effectiveness of the cross-attention operator in balancing the trade-off between linear and non-linear similarity functions.

\input{tables/ab_simfunc}

\input{tables/ab_encoder}

\noindent\textbf{Generalization to Diverse Cross-modal Settings.}
To validate the generalizability of the proposed method across a flexible number of feature encoders, we experiment with different settings: unimodal, paired (video-text, audio-text), and unified embedding space (image-audio-text).
In the unimodal setting, we use DINO~\cite{caron2021emerging} for the visual encoder, AudioMAE~\cite{huang2022amae} for the audio encoder, and LLaMa~\cite{touvron2023llama} for the text encoder.
In the paired setting, we use  
X-CLIP~\cite{Ma2022XCLIP} pre-trained with video-text objectives for the visual encoder, and CLAP~\cite{laionclap2023} pre-trained with audio-text objectives for the audio and text encoder, respectively.
In the unified embedding setting, we apply ImageBind~\cite{girdhar2023imagebind} pre-trained with image-audio-text alignment in the same embedding space as the encoder for all three modalities. 
The quantitative results on the VGGSound-GZSL benchmark are compared in Table~\ref{tab: ab_encoder}.
As observed, the unimodal setting, which uses transformers with strong unimodal features, significantly outperforms the vanilla baseline that uses convolution networks, where \method achieves performance gains of 6.69@Seen, 4.64@Unseen, 5.77@Harmonic Mean, and 4.81@ZSL using class embedding optimization.
In the paired setting, the proposed method further increases the results by 9.33@Seen, 6.93@Unseen, 7.98@Harmonic Mean, and 7.331@ZSL, benefiting from the text-audio correspondence and text-video alignment pre-trained on large-scale data.  
In the unified embedding setting using ImageBind~\cite{girdhar2023imagebind} embeddings, we achieve the best results across all metrics compared to other settings.
This might be because text embeddings with unified training on the other two modalities strengthen the alignment for audio-visual zero-shot learning.
Furthermore, when class embedding optimization is added to all settings, we improve the baseline based on the initial text embeddings, further demonstrating the effectiveness of the proposed class embedding optimization in generating text representations for AVZSL.

\input{tables/ab_cls}

\input{figs/exp_vis}

\noindent\textbf{Qualitative Comparisons with and without Optimization.}
To provide insights into why the proposed loss terms improve the separability of class embeddings while keeping semantics, we conduct qualitative studies. These include comparing the visualization of the class embedding space with and without optimization.
The comparisons of class embeddings with and without optimization are shown in Table~\ref{tab: exp_cls}.
We report the base class and identify 1) the nearest neighbor (NN) class and 2) the distance (D) to the nearest neighbor. 
We observe larger distances and similar NN classes. The confusion matrices for class-wise performances of the model are reported in Fig.~\ref{fig: vis_confusion}.

\input{tables/ab_alpha_m}

\noindent\textbf{Impact of Hyper-parameters $\alpha$ and Margin $m$.}
To investigate the impact of $\alpha$ and $m$ on the performance thoroughly, we set $\alpha$ from $\{0.1,0.5,0.9\}$ and $m$ from $\{0,0.5,1\}$.
The quantitative comparisons on VGGSound-GZSL are reported in Table~\ref{tab: ab_alpha_m}.
As can be seen, with the increase of $\alpha$ from $0.1$ to $0.5$, the proposed approach achieves better results regarding all metrics.
However, when $\alpha$ is increased to $0.9$, the performance decreases a lot, which might be caused by not much semantics preservation during training. 
Meanwhile, with the decrease of $m$, we observe a clearly dropping trend in our metrics. 

\noindent\textbf{Ablating Distance Functions for Eq.~\ref{eq:sep}-\ref{eq:sem_prox}.}
Distance functions in Eq.~\ref{eq:sep}-\ref{eq:sem_prox} is crucial for us to maximize separation while preserving semantics. 
We experimented with different distance functions, $\ell_2$ and $\ell_1$ distances for Eq.~\ref{eq:sep}-\ref{eq:sem_prox}. 
Note that while, Eq.~\ref{eq:sep}-\ref{eq:sem_prox} show $\ell_2$ distances in the paper, the embeddings are normalized.
The quantitative results are shown in Table~\ref{tab: ab_distance}. 
We can observe that using the cosine distance function performed the best regarding all metrics.
However, when optimizing $\ell_1$ distance functions, we achieve the worst results in terms of all metrics.
These ablation results highlight the importance of designing appropriate distance functions in maximizing separation while preserving the semantics of text class embeddings.

%% file: tables/ab_component.tex
\begin{table}[t]
   \centering
   \begin{minipage}{0.43\linewidth}
       \centering
       \caption{Ablation studies on Class Embedding Optimization and Audio-Visual-Language Alignment.}
   \label{tab: exp_ablation}
	\resizebox{\linewidth}{!}{
		\begin{tabular}{cccccc}
			\toprule
    \bf CEP & \bf AVLA & \bf Seen & \bf Unseen & \bf Harmonic Mean & \bf ZSL \\
   \midrule
    \textcolor{ForestGreen}{\xmark} & \textcolor{ForestGreen}{\xmark} & 12.63 & 6.72 & 8.77 & 7.41 \\
    \cmark & \textcolor{ForestGreen}{\xmark} & 15.35	 & 7.92 & 10.45	& 8.36 \\
    \textcolor{ForestGreen}{\xmark} & \cmark & 14.67	 & 7.23 & 9.69 & 7.91 \\
    \rowcolor{blue!10}
    \cmark & \cmark & \bf 17.26	 & \bf 8.68 & \bf 11.55 & \bf 8.92 \\
   \bottomrule
	\end{tabular}}
   \end{minipage}\hfill
   \begin{minipage}{0.55\linewidth}
      \centering
      \caption{Ablation studies on optimization objectives in Class Embedding Optimization.}
   \label{tab: ab_textloss}
	\resizebox{\linewidth}{!}{
		\begin{tabular}{cccccc}
			\toprule
    $\boldsymbol{\mathcal{L}_{sem}}$ & $\boldsymbol{\mathcal{L}_{sep}}$ & \bf Seen & \bf Unseen & \bf Harmonic Mean & \bf ZSL \\
   \midrule
    \textcolor{ForestGreen}{\xmark} & \textcolor{ForestGreen}{\xmark} & 14.67 & 7.23 & 9.32 & 7.91 \\
    $\mathcal{L}^{\textit{prox}}_{sem}$ & \textcolor{ForestGreen}{\xmark} & 15.75 & 7.83 & 10.46 & 8.26 \\
    \textcolor{ForestGreen}{\xmark} & \cmark  & 16.13 & 8.17 & 10.85 & 8.45 \\
    $\mathcal{L}^{\textit{prox}}_{sem}$ & \cmark & 17.26	 & 8.68 & 11.55 &  8.92 \\
    \rowcolor{blue!10}
    $\mathcal{L}_{sem}^{\textit{rank}}$ & \cmark & \bf 18.05 & \bf 10.28 & \bf 13.10	& \bf 10.36 \\
   \bottomrule
	\end{tabular}}
   \end{minipage}
\end{table}

%% file: tables/ab_simfunc.tex
\begin{table}[t]
	%\normalem
	\renewcommand\tabcolsep{3.0pt}
	\centering
        \caption{Exploration studies on non-linear similarity functions $\phi(\cdot)$.}
   \label{tab: ab_simfunc}
	\resizebox{0.45\linewidth}{!}{
		\begin{tabular}{ccccc}
			\toprule
     $\boldsymbol{\phi(\cdot)}$ & \bf Seen & \bf Unseen & \bf Harmonic Mean & \bf ZSL \\
   \midrule
    Cosine-Similarity & 13.97	& 8.59 & 10.64 & 8.67 \\
    Linear & 16.52	& 8.52 & 11.24 & 8.55 \\
    MLP & 17.03	& 8.46 & 11.30 & 8.49 \\
    \rowcolor{blue!10}
    Cross-Attention & \bf 17.26	 & \bf 8.68 & \bf 11.55 & \bf 8.92 \\
   \bottomrule
	\end{tabular}}
\end{table}

%% file: tables/ab_encoder.tex
\begin{table*}[t]
	%\normalem
    \renewcommand\tabcolsep{3.0pt}
    \centering
    \caption{Exploration studies on diverse cross-modal settings. CEP denotes Class Embedding Optimization.}
    \label{tab: ab_encoder}
    \resizebox{0.55\linewidth}{!}{
    \begin{tabular}{ccccllll}
    \toprule
    \bf \thead{Text\\Encoder} & \bf \thead{Audio\\Encoder} & \bf \thead{Visual\\Encoder} & \bf \thead{CEP} & \bf Seen & \bf Unseen & \bf \thead{Harmonic\\Mean} & \bf ZSL \\
    \midrule
    \multirow{2}{*}{word2vec} & \multirow{2}{*}{VGGish} & \multirow{2}{*}{C3D} & \textcolor{ForestGreen}{\xmark} &  12.63 & 6.72 & 8.77 & 7.41 \\
    
     & & & \cellcolor{blue!10}\cmark &  \cellcolor{blue!10}\bf 17.26 & \cellcolor{blue!10}\bf 8.68 & \cellcolor{blue!10}\bf 11.55 & \cellcolor{blue!10}\bf 8.92 \\\midrule
    \multirow{2}{*}{LLaMa} & \multirow{2}{*}{AudioMAE} & \multirow{2}{*}{DINO} & \textcolor{ForestGreen}{\xmark} & 18.35 & 10.63 & 13.46 & 12.26 \\
     &  &  & \cellcolor{blue!10}\cmark & \cellcolor{blue!10}\bf 23.95 & \cellcolor{blue!10}\bf 13.32 & \cellcolor{blue!10}\bf 17.12 & \cellcolor{blue!10}\bf 13.73 \\\midrule
     \multirow{2}{*}{CLAP} & \multirow{2}{*}{CLAP} & \multirow{2}{*}{X-CLIP} & \textcolor{ForestGreen}{\xmark} & 27.79 & 14.01 & 18.63 & 17.39 \\
    & & & \cellcolor{blue!10}\cmark & \cellcolor{blue!10}\bf 33.28 & \cellcolor{blue!10}\bf 20.15 & \cellcolor{blue!10}\bf 25.10 & \cellcolor{blue!10}\bf 21.06 \\\midrule
    \multirow{2}{*}{ImageBind} & \multirow{2}{*}{ImageBind} & \multirow{2}{*}{ImageBind} & \textcolor{ForestGreen}{\xmark} & 28.32 & 14.68 & 19.34 & 19.87 \\
    & & & \cellcolor{blue!10}\cmark & \cellcolor{blue!10}\bf 35.56 & \cellcolor{blue!10}\bf 21.63 & \cellcolor{blue!10}\bf 26.90 & \cellcolor{blue!10}\bf 22.35 \\
    \bottomrule
    \end{tabular}}
\end{table*}

%% file: tables/ab_cls.tex
\begin{table}[t]
	%\normalem
	\renewcommand\tabcolsep{6.0pt}
	\centering
        \caption{Comparison of class embeddings w/o and w optimization.}
   \label{tab: exp_cls}
	\resizebox{0.8\linewidth}{!}{
		\begin{tabular}{ccccc}
			\toprule
    \bf base class & \bf NN (w/o opt) & \bf D (w/o opt) & \bf NN (w opt) & \bf D (w opt) \\
   \midrule
  alligators,   crocodiles hissing & church bell ringing               & 0.455 & playing marimba, xylophone     & \cellcolor{blue!10}\bf 0.498 \\
barn swallow calling             & black capped chickadee calling    & 0.307 & black capped chickadee calling & \cellcolor{blue!10}\bf 0.465 \\
basketball bounce                & playing volleyball                & 0.324 & bouncing on trampoline         & \cellcolor{blue!10}\bf 0.394 \\
bee, wasp, etc. buzzing          & motorboat, speedboat acceleration & 0.322 & fly, housefly buzzing          & \cellcolor{blue!10}\bf 0.333 \\
bird squawking                   & pheasant crowing                  & 0.400 & duck quacking                  & \cellcolor{blue!10}\bf 0.496 \\
black capped chickadee calling   & barn swallow calling              & 0.307 & magpie calling                 & \cellcolor{blue!10}\bf 0.436 \\
bouncing on trampoline           & basketball bounce                 & 0.416 & basketball bounce              & \cellcolor{blue!10}\bf 0.494 \\
bowling impact                   & playing table tennis              & 0.410 & striking bowling               & \cellcolor{blue!10}\bf 0.467 \\
bull bellowing                   & eagle screaming                   & 0.449 & dog growling                   & \cellcolor{blue!10}\bf 0.532 \\
canary calling                   & whale calling                     & 0.403 & black capped chickadee calling & \cellcolor{blue!10}\bf 0.513 \\
   \bottomrule
	\end{tabular}}
\end{table}

%% file: figs/exp_vis.tex
\begin{figure}[t]
\centering 
% \fbox{\rule{0pt}{1in}
% \rule{0.8\linewidth}{0pt}}
\includegraphics[width=0.9\linewidth]{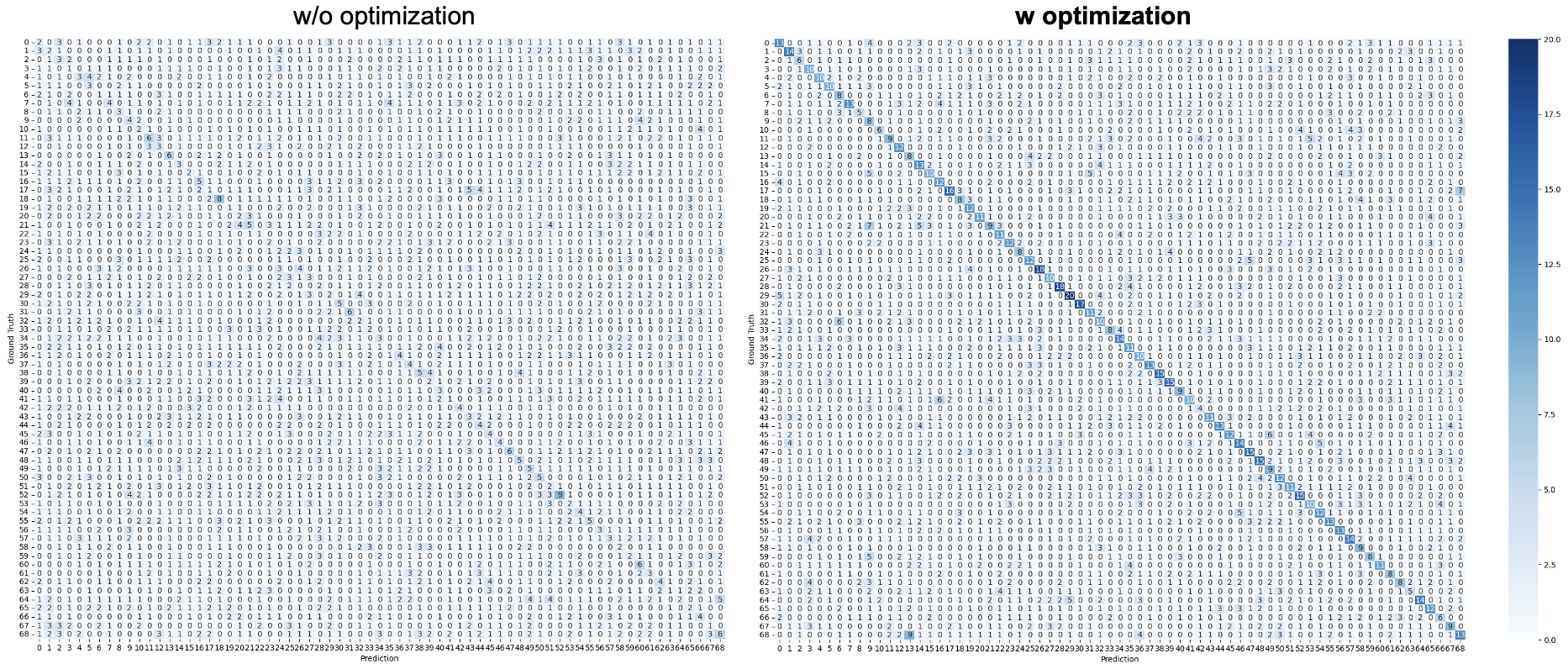}
\caption{Confusion matrices of zero-shot predictions on VGGSound-GZSL using model with and without class embedding optimization.}
\label{fig: vis_confusion}
\end{figure}

%% file: tables/ab_alpha_m.tex
\begin{table}[t]
   \centering
   \begin{minipage}{0.4\linewidth}
       \centering
       \caption{Study of $\alpha$ and margin $m$ on VGGSound-GZSL.\label{tab: ab_alpha_m}}
       \resizebox{\linewidth}{!}{
       \begin{tabular}{cccccc}
          \toprule
          $\boldsymbol{\alpha}$ & $\boldsymbol{m}$ & \bf Seen & \bf Unseen & \bf Harmonic Mean & \bf ZSL \\
          \midrule
          0.5 & \multirow{3}{*}{1} & \cellcolor{blue!10}\bf 18.05 & \cellcolor{blue!10}\bf 10.28 & \cellcolor{blue!10}\bf 13.10	& \cellcolor{blue!10}\bf 10.36 \\
          0.1 & & 16.35  & 8.39 & 11.09 & 8.58 \\
          0.9 & & 16.26	& 8.73 & 11.36 & 8.92 \\\hline
          \multirow{2}{*}{0.5} & 0 & 17.82 & 9.71 & 12.57 & 9.67 \\
          & 0.5 & 17.67 & 9.85 & 12.65 & 9.95 \\
          \bottomrule
       \end{tabular}}
   \end{minipage}\hfill
   \begin{minipage}{0.55\linewidth}
      \centering
      \caption{Study of distance functions on VGGSound-GZSL. \label{tab: ab_distance}}
      \resizebox{\linewidth}{!}{
      \begin{tabular}{ccccc}
         \toprule
         \bf Distance & \bf Seen & \bf Unseen & \bf Harmonic Mean & \bf ZSL \\
         \midrule
         cosine & \cellcolor{blue!10}\bf 18.05 & \cellcolor{blue!10}\bf 10.28 & \cellcolor{blue!10}\bf 13.10	& \cellcolor{blue!10}\bf 10.36 \\
         $\ell_2$ & 17.58 & 9.32 & 12.18 & 9.27 \\
         $\ell_1$ & 17.35 & 8.78 & 11.66 & 8.85 \\
         \bottomrule
      \end{tabular}}
   \end{minipage}
\end{table}

%% file: latex/6-conclusion.tex
\section{Conclusion}

In this work, we present a novel and effective framework, \method, that aligns audio-visual representations with optimized text embeddings to achieve audio-visual generalized zero-shot learning. 
We leverage differential optimization to learn better-separated textual representations with maximal separability and semantic preservation.
Additionally, we introduce supervised audio-visual language alignment to learn the correspondence between audio-visual features and textual embeddings, which captures the cross-modal dynamics with class labels as prompts. Experimental results on VGGSound-GZSL, UCF-GZSL, and ActivityNet-GZSL datasets demonstrate our method's superiority against previous baselines. Extensive ablation studies also validate the importance of class embedding optimization and supervised audio-visual language alignment for audio-visual generalized zero-shot learning. We also conduct comprehensive experiments to demonstrate the effectiveness of non-linear similarity functions and generalization to diverse cross-modal settings.

\noindent\textbf{Broader Impact.}
This work makes substantial progress in audio-visual generalized zero-shot learning. 
Our method is novel in its simplicity, demonstrating that a simple supervised contrastive objective, without "bells and whistles", outperforms complex alignment strategies. 
Without our work, progress in this area would likely be driven by increasingly complex methods, which are difficult to implement, reproduce and compare. 
Beyond the strong results and simplicity of our method, the class embedding optimization procedure is also technically novel to maximize semantic separation while keeping semantics preserved.

%% file: main.bbl
\begin{thebibliography}{10}
\providecommand{\url}[1]{\texttt{#1}}
\providecommand{\urlprefix}{URL }
\providecommand{\doi}[1]{https://doi.org/#1}

\bibitem{Arandjelovic2017look}
Arandjelovic, R., Zisserman, A.: Look, listen and learn. In: Proceedings of the IEEE International Conference on Computer Vision (ICCV). pp. 609--617 (2017)

\bibitem{aytar2016soundnet}
Aytar, Y., Vondrick, C., Torralba, A.: Soundnet: Learning sound representations from unlabeled video. In: Proceedings of Advances in Neural Information Processing Systems (NeurIPS) (2016)

\bibitem{caron2021emerging}
Caron, M., Touvron, H., Misra, I., J\'egou, H., Mairal, J., Bojanowski, P., Joulin, A.: Emerging properties in self-supervised vision transformers. In: Proceedings of the International Conference on Computer Vision (ICCV) (2021)

\bibitem{Chen2020SoundSpacesAN}
Chen, C., Jain, U., Schissler, C., Gar{\'i}, S.V.A., Al-Halah, Z., Ithapu, V.K., Robinson, P., Grauman, K.: Soundspaces: Audio-visual navigation in 3d environments. In: Proceedings of European Conference on Computer Vision (ECCV). pp. 17--36 (2020)

\bibitem{chen2021waypoints}
Chen, C., Majumder, S., Ziad, A.H., Gao, R., Kumar~Ramakrishnan, S., Grauman, K.: Learning to set waypoints for audio-visual navigation. In: Proceedings of International Conference on Learning Representations (ICLR) (2021)

\bibitem{chen22soundspaces2}
Chen, C., Schissler, C., Garg, S., Kobernik, P., Clegg, A., Calamia, P., Batra, D., Robinson, P.W., Grauman, K.: Soundspaces 2.0: A simulation platform for visual-acoustic learning. In: Proceedings of Advances in Neural Information Processing Systems (NeurIPS) Datasets and Benchmarks Track (2022)

\bibitem{chen2020vggsound}
Chen, H., Xie, W., Vedaldi, A., Zisserman, A.: Vggsound: A large-scale audio-visual dataset. In: ICASSP 2020-2020 IEEE International Conference on Acoustics, Speech and Signal Processing (ICASSP). pp. 721--725. IEEE (2020)

\bibitem{ephrat2018looking}
Ephrat, A., Mosseri, I., Lang, O., Dekel, T., Wilson, K., Hassidim, A., Freeman, W.T., Rubinstein, M.: Looking to listen at the cocktail party: A speaker-independent audio-visual model for speech separation. arXiv preprint arXiv:1804.03619  (2018)

\bibitem{Fayek2021large}
Fayek, H.M., Kumar, A.: Large scale audiovisual learning of sounds with weakly labeled data. In: Proceedings of the Twenty-Ninth International Joint Conference on Artificial Intelligence (2021)

\bibitem{Gan2020music}
Gan, C., Huang, D., Zhao, H., Tenenbaum, J.B., Torralba, A.: Music gesture for visual sound separation. In: IEEE/CVF Conference on Computer Vision and Pattern Recognition (CVPR). pp. 10478--10487 (2020)

\bibitem{gao20192.5D}
Gao, R., Grauman, K.: 2.5d visual sound. In: Proceedings of the IEEE/CVF Conference on Computer Vision and Pattern Recognition (CVPR). pp. 324--333 (2019)

\bibitem{girdhar2023imagebind}
Girdhar, R., El-Nouby, A., Liu, Z., Singh, M., Alwala, K.V., Joulin, A., Misra, I.: Imagebind: One embedding space to bind them all. In: Proceedings of the IEEE/CVF Conference on Computer Vision and Pattern Recognition (CVPR) (2023)

\bibitem{heilbron2015activitynet}
Heilbron, F.C., Escorcia, V., Ghanem, B., Niebles, J.C.: Activitynet: A large-scale video benchmark for human activity understanding. In: 2015 IEEE Conference on Computer Vision and Pattern Recognition (CVPR). pp. 961--970 (2015)

\bibitem{hershey2001audio}
Hershey, J., Casey, M.: Audio-visual sound separation via hidden markov models. Advances in Neural Information Processing Systems  \textbf{14} (2001)

\bibitem{hershey2017cnn}
Hershey, S., Chaudhuri, S., Ellis, D.P.W., Gemmeke, J.F., Jansen, A., Moore, R.C., Plakal, M., Platt, D., Saurous, R.A., Seybold, B., Slaney, M., Weiss, R.J., Wilson, K.: Cnn architectures for large-scale audio classification. In: IEEE International Conference on Acoustics, Speech and Signal Processing (ICASSP) (2017)

\bibitem{hong2023hyperbolic}
Hong, J., Hayder, Z., Han, J., Fang, P., Harandi, M., Petersson, L.: Hyperbolic audio-visual zero-shot learning. In: Proceedings of the International Conference on Computer Vision (ICCV) (2023)

\bibitem{hu2019deep}
Hu, D., Nie, F., Li, X.: Deep multimodal clustering for unsupervised audiovisual learning. In: Proceedings of the IEEE Conference on Computer Vision and Pattern Recognition (CVPR). pp. 9248--9257 (2019)

\bibitem{huang2022amae}
Huang, P.Y., Xu, H., Li, J., Baevski, A., Auli, M., Galuba, W., Metze, F., Feichtenhofer, C.: Masked autoencoders that listen. In: Proceedings of Advances In Neural Information Processing Systems (NeurIPS) (2022)

\bibitem{jaegle2021perceiver}
Jaegle, A., Gimeno, F., Brock, A., Zisserman, A., Vinyals, O., Carreira, J.: Perceiver: General perception with iterative attention. In: Proceedings of the International Conference on Machine Learning (ICML) (2021)

\bibitem{kingma2014adam}
Kingma, D.P., Ba, J.: {Adam}: A method for stochastic optimization. arXiv preprint arXiv:1412.6980  (2014)

\bibitem{korbar2018cooperative}
Korbar, B., Tran, D., Torresani, L.: Cooperative learning of audio and video models from self-supervised synchronization. In: Proceedings of Advances in Neural Information Processing Systems (NeurIPS) (2018)

\bibitem{lin2019dual}
Lin, Y.B., Li, Y.J., Wang, Y.C.F.: Dual-modality seq2seq network for audio-visual event localization. In: IEEE International Conference on Acoustics, Speech and Signal Processing (ICASSP). pp. 2002--2006 (2019)

\bibitem{lin2021exploring}
Lin, Y.B., Tseng, H.Y., Lee, H.Y., Lin, Y.Y., Yang, M.H.: Exploring cross-video and cross-modality signals for weakly-supervised audio-visual video parsing. In: Proceedings of Advances in Neural Information Processing Systems (NeurIPS) (2021)

\bibitem{lin2020audiovisual}
Lin, Y.B., Wang, Y.C.F.: Audiovisual transformer with instance attention for audio-visual event localization. In: Proceedings of the Asian Conference on Computer Vision (ACCV) (2020)

\bibitem{Ma2022XCLIP}
Ma, Y., Xu, G., Sun, X., Yan, M., Zhang, J., Ji, R.: {X-CLIP:}: End-to-end multi-grained contrastive learning for video-text retrieval. arXiv preprint arXiv:2207.07285  (2022)

\bibitem{Mazumder2021avgzslnet}
Mazumder, P., Sing, P., Kumar~Parida, K., Namboodiri, V.P.: Avgzslnet: Audio-visual generalized zero-shot learning by reconstructing label features from multi-modal embeddings. In: Proceedings of 2021 IEEE Winter Conference on Applications of Computer Vision (WACV). pp. 3089--3098 (2021)

\bibitem{mercea2022tcaf}
Mercea, O.B., Hummel, T., Koepke, A.S., Akata, Z.: Temporal and cross-modal attention for audio-visual zero-shot learning. In: Proceedings of European Conference on Computer Vision (ECCV) (2022)

\bibitem{mercea2022avca}
Mercea, O.B., Riesch, L., Koepke, A.S., Akata, Z.: Audio-visual generalised zero-shot learning with cross-modal attention and language. In: Proceedings of the IEEE/CVF Conference on Computer Vision and Pattern Recognition (CVPR). pp. 10553--10563 (2022)

\bibitem{mikolov2013efficient}
Mikolov, T., Chen, K., Corrado, G., Dean, J.: Efficient estimation of word representations in vector space. In: Proceedings of International Conference on Learning Representations (ICLR) (2013)

\bibitem{mo2022benchmarking}
Mo, S., Morgado, P.: Benchmarking weakly-supervised audio-visual sound localization. In: European Conference on Computer Vision (ECCV) Workshop (2022)

\bibitem{mo2022SLAVC}
Mo, S., Morgado, P.: A closer look at weakly-supervised audio-visual source localization. In: Proceedings of Advances in Neural Information Processing Systems (NeurIPS) (2022)

\bibitem{mo2023oneavm}
Mo, S., Morgado, P.: A unified audio-visual learning framework for localization, separation, and recognition. In: Proceedings of the International Conference on Machine Learning (ICML) (2023)

\bibitem{mo2023deepavfusion}
Mo, S., Morgado, P.: Unveiling the power of audio-visual early fusion transformers with dense interactions through masked modeling. arXiv preprint arXiv:2312.01017  (2023)

\bibitem{mo2023classincremental}
Mo, S., Pian, W., Tian, Y.: Class-incremental grouping network for continual audio-visual learning. arXiv preprint arXiv:2309.05281  (2023)

\bibitem{mo2023diffava}
Mo, S., Shi, J., Tian, Y.: {DiffAVA}: Personalized text-to-audio generation with visual alignment. arXiv preprint arXiv:2305.12903  (2023)

\bibitem{mo2024texttoaudio}
Mo, S., Shi, J., Tian, Y.: Text-to-audio generation synchronized with videos. arXiv preprint arXiv:2403.07938  (2024)

\bibitem{mo2022multimodal}
Mo, S., Tian, Y.: Multi-modal grouping network for weakly-supervised audio-visual video parsing. In: Proceedings of Advances in Neural Information Processing Systems (NeurIPS) (2022)

\bibitem{mo2022semantic}
Mo, S., Tian, Y.: Semantic-aware multi-modal grouping for weakly-supervised audio-visual video parsing. In: European Conference on Computer Vision (ECCV) Workshop (2022)

\bibitem{mo2023audiovisual}
Mo, S., Tian, Y.: Audio-visual grouping network for sound localization from mixtures. arXiv preprint arXiv:2303.17056  (2023)

\bibitem{mo2023avsam}
Mo, S., Tian, Y.: {AV-SAM}: Segment anything model meets audio-visual localization and segmentation. arXiv preprint arXiv:2305.01836  (2023)

\bibitem{mo2024semantic}
Mo, S., Tian, Y.: Semantic grouping network for audio source separation. arXiv preprint arXiv:2407.03736  (2024)

\bibitem{Morgado2020learning}
Morgado, P., Li, Y., Vasconcelos, N.: Learning representations from audio-visual spatial alignment. In: Proceedings of Advances in Neural Information Processing Systems (NeurIPS). pp. 4733--4744 (2020)

\bibitem{MorgadoProxyHashing}
Morgado, P., Li, Y., Costa~Pereira, J., Saberian, M., Vasconcelos, N.: Deep hashing with hash-consistent large margin proxy embeddings. International Journal of Computer Vision  (2020)

\bibitem{Morgado2021robust}
Morgado, P., Misra, I., Vasconcelos, N.: Robust audio-visual instance discrimination. In: Proceedings of the IEEE/CVF Conference on Computer Vision and Pattern Recognition (CVPR). pp. 12934--12945 (2021)

\bibitem{Morgado2018selfsupervised}
Morgado, P., Vasconcelos, N., Langlois, T., Wang, O.: Self-supervised generation of spatial audio for 360 video. In: Proceedings of Advances in Neural Information Processing Systems (NeurIPS) (2018)

\bibitem{Morgado2021audio}
Morgado, P., Vasconcelos, N., Misra, I.: Audio-visual instance discrimination with cross-modal agreement. In: Proceedings of the IEEE/CVF Conference on Computer Vision and Pattern Recognition (CVPR). pp. 12475--12486 (June 2021)

\bibitem{owens2016ambient}
Owens, A., Wu, J., McDermott, J.H., Freeman, W.T., Torralba, A.: Ambient sound provides supervision for visual learning. In: Proceedings of the European Conference on Computer Vision (ECCV). pp. 801--816 (2016)

\bibitem{Parida2020cjme}
Parida, K.K., Matiyali, N., Guha, T., Sharma, G.: Coordinated joint multimodal embeddings for generalized audio-visual zero-shot classification and retrieval of videos. In: Proceedings of 2020 IEEE Winter Conference on Applications of Computer Vision (WACV). pp. 3240--3249 (2020)

\bibitem{pian2023audiovisual}
Pian, W., Mo, S., Guo, Y., Tian, Y.: Audio-visual class-incremental learning. arXiv preprint arXiv:2308.11073  (2023)

\bibitem{Senocak2018learning}
Senocak, A., Oh, T.H., Kim, J., Yang, M.H., Kweon, I.S.: Learning to localize sound source in visual scenes. In: Proceedings of the IEEE Conference on Computer Vision and Pattern Recognition (CVPR). pp. 4358--4366 (2018)

\bibitem{soomro2012ucf101}
Soomro, K., Zamir, A.R., Shah, M.: Ucf101: A dataset of 101 human actions classes from videos in the wild. arXiv preprint arXiv:1212.0402  (2012)

\bibitem{tammes1930origin}
Tammes, P.M.L.: On the origin of number and arrangement of the places of exit on the surface of pollen-grains. Recueil des travaux botaniques n{\'e}erlandais  \textbf{27}(1),  1--84 (1930)

\bibitem{tian2020avvp}
Tian, Y., Li, D., Xu, C.: Unified multisensory perception: Weakly-supervised audio-visual video parsing. In: Proceedings of European Conference on Computer Vision (ECCV). p. 436–454 (2020)

\bibitem{tian2018ave}
Tian, Y., Shi, J., Li, B., Duan, Z., Xu, C.: Audio-visual event localization in unconstrained videos. In: Proceedings of European Conference on Computer Vision (ECCV) (2018)

\bibitem{touvron2023llama}
Touvron, H., Martin, L., Stone, K., Albert, P., Almahairi, A., Babaei, Y., Bashlykov, N., Batra, S., Bhargava, P., Bhosale, S., Bikel, D., Blecher, L., Ferrer, C.C., Chen, M., Cucurull, G., Esiobu, D., Fernandes, J., Fu, J., Fu, W., Fuller, B., Gao, C., Goswami, V., Goyal, N., Hartshorn, A., Hosseini, S., Hou, R., Inan, H., Kardas, M., Kerkez, V., Khabsa, M., Kloumann, I., Korenev, A., Koura, P.S., Lachaux, M.A., Lavril, T., Lee, J., Liskovich, D., Lu, Y., Mao, Y., Martinet, X., Mihaylov, T., Mishra, P., Molybog, I., Nie, Y., Poulton, A., Reizenstein, J., Rungta, R., Saladi, K., Schelten, A., Silva, R., Smith, E.M., Subramanian, R., Tan, X.E., Tang, B., Taylor, R., Williams, A., Kuan, J.X., Xu, P., Yan, Z., Zarov, I., Zhang, Y., Fan, A., Kambadur, M., Narang, S., Rodriguez, A., Stojnic, R., Edunov, S., Scialom, T.: Llama 2: Open foundation and fine-tuned chat models. arXiv preprint arXiv:2307.09288  (2023)

\bibitem{tran2015learning}
Tran, D., Bourdev, L., Fergus, R., Torresani, L., Paluri, M.: Learning spatiotemporal features with 3d convolutional networks. In: Proceedings of the IEEE International Conference on Computer Vision (ICCV) (2015)

\bibitem{wu2021explore}
Wu, Y., Yang, Y.: Exploring heterogeneous clues for weakly-supervised audio-visual video parsing. In: Proceedings of the IEEE Conference on Computer Vision and Pattern Recognition (CVPR). pp. 1326--1335 (2021)

\bibitem{wu2019dual}
Wu, Y., Zhu, L., Yan, Y., Yang, Y.: Dual attention matching for audio-visual event localization. In: Proceedings of the IEEE International Conference on Computer Vision (ICCV). pp. 6291--6299 (2019)

\bibitem{laionclap2023}
Wu, Y., Chen, K., Zhang, T., Hui, Y., Berg-Kirkpatrick, T., Dubnov, S.: Large-scale contrastive language-audio pretraining with feature fusion and keyword-to-caption augmentation. In: IEEE International Conference on Acoustics, Speech and Signal Processing, ICASSP (2023)

\bibitem{zhang2024audiosynchronized}
Zhang, L., Mo, S., Zhang, Y., Morgado, P.: Audio-synchronized visual animation. arXiv preprint arXiv:2403.05659  (2024)

\bibitem{zhao2019the}
Zhao, H., Gan, C., Ma, W.C., Torralba, A.: The sound of motions. In: Proceedings of the IEEE/CVF International Conference on Computer Vision (ICCV). pp. 1735--1744 (2019)

\bibitem{zhao2018the}
Zhao, H., Gan, C., Rouditchenko, A., Vondrick, C., McDermott, J., Torralba, A.: The sound of pixels. In: Proceedings of the European Conference on Computer Vision (ECCV). pp. 570--586 (2018)

\end{thebibliography}
